\documentclass{styles/svproc}
\usepackage[table]{xcolor}
\usepackage{tabularx}
\usepackage{algorithmic}
\usepackage{units}
\usepackage[english]{babel}

\usepackage{color}
\usepackage[flushleft]{threeparttable}

\usepackage{comment}
\usepackage{datetime}

\usepackage{epsfig}         
\usepackage{graphics}       
\usepackage{graphicx}

\usepackage{enumerate}
\usepackage{enumitem}

\usepackage{xspace}

\usepackage{tabularx}

\usepackage[colorlinks, pagebackref=false]{hyperref} 

\definecolor{mygreen}{rgb}{0,0.5,0}
\definecolor{myblue}{rgb}{0,0,0.8} 
\PassOptionsToPackage{unicode}{hyperref}
\hypersetup
{ colorlinks=true,
	linkcolor=myblue,
	citecolor=mygreen,
	filecolor=magenta,
	urlcolor=myblue,
	linktoc=page,
}

\usepackage{bibunits} 

\usepackage{refcount}
\usepackage{sepfootnotes} 

\usepackage{breakcites} 

\usepackage{amssymb}
\usepackage{amsfonts}
\usepackage{amsmath}
\usepackage{mathtools}

\usepackage{graphicx}
\usepackage{float}

\usepackage{url}
\usepackage{orcidlink}

\usepackage{hyperref}

\newcommand{\shortlink}[1]{\href{https://www.\detokenize{#1}}{\texttt{#1}}}

\DeclareMathSymbol{\shortminus}{\mathbin}{AMSa}{"39}

\DeclareUnicodeCharacter{2212}{-}

\usepackage{url}

\usepackage[official]{eurosym}

\begin{document}
\mainmatter              
\title{Advances on Affordable Hardware Platforms for \\ Human Demonstration Acquisition \\ in Agricultural Applications}
\titlerunning{Adv. on Aff. HW Platforms for Human Demo. Acquisition in Agro App.}  
%
\author{Alberto San-Miguel-Tello \orcidlink{0000-0001-5547-2564}
\and 
Gennaro Scarati \orcidlink{0009-0000-5255-0635}
\and 
Alejandro Hernández \orcidlink{0009-0007-5270-6573}
\and \\
Mario Cavero-Vidal\orcidlink{0009-0008-5877-4283}
\and
Aakash Maroti\orcidlink{0009-0009-4208-9107}
\and
Néstor García \orcidlink{0000-0003-3782-1745}
}
%
\authorrunning{A. San-Miguel-Tello et al.} 
%
\tocauthor{Alberto San-Miguel-Tello, Gennaro Scarati, Alejandro Hernández, Mario Cavero-Vidal, Aakash Maroti and Néstor García}
\institute{Eurecat, Centre Tecnològic de Catalunya, Robotics and Automation Unit, Barcelona, 08290, Spain. {\tt\scriptsize \{alberto.sanmiguel, gennaro.scarati, alejandro.gonzalez, mario.cavero, aakash.maroti, nestor.garcia\}@eurecat.org}}

\maketitle              

\vspace{-1mm}
\begin{abstract}
This paper presents advances on the Universal Manipulation Interface (UMI)~\cite{chi_universal_2024}, a low-cost hand-held gripper for robot Learning from Demonstration (LfD), for complex in-the-wild scenarios found in agricultural settings. The focus is on improving the acquisition of suitable samples with minimal additional setup. Firstly, idle times and user's cognitive load are reduced through the extraction of individual samples from a continuous demonstration considering task events. Secondly, reliability on the generation of task sample's trajectories is increased through the combination on-board inertial measurements and external visual marker localization usage using Extended Kalman Filtering (EKF). Results are presented for a fruit harvesting task, outperforming the default pipeline.
\end{abstract}
\section{Introduction}
\vspace{-2mm}
Despite the recent advances on robot autonomy and cognition, many agricultural activities still require human-lead manual activities, compared to other domains like the industrial one~\cite{wakchaure_application_2023}, despite the associated high costs and risks. Most of the existing solutions aim at autonomous navigation of mobile platforms with standard tools, e.g. plough, sprayer or combine harvester, usually extended with other methods to perform decision-making operations~\cite{gao_uav_2021}. Remaining solutions focus on monitoring processes using Computer Vision techniques, e.g. detect weeds~\cite{hall_weed_2017}. One of the tasks with the lowest degrees of automation is fruit harvesting, where solutions leverage on robotic manipulators to perform it, sometimes in combination with end-effector design to ensure damage-free grasping~\cite{gunderman_tendon_2022}, and advance on the required monitoring techniques, e.g. to provide the 3D pose of the fruit for picking movement planning~\cite{tao_automatic_2017}. But for motion generation, these strategies rely on invariant approach-grasping-retreat strategies, which differs from the complex motions that human experts perform and have been proved to play an important role in fruit quality~\cite{li_apple_2016}. Learning from Demonstration (LfD) approaches could provide an alternative to this paradigm, but, up the best of our knowledge, there hasn't been any attempt to introduce them on this application.

\begin{figure}[ht]
    \centering
    \includegraphics[width=1\textwidth]{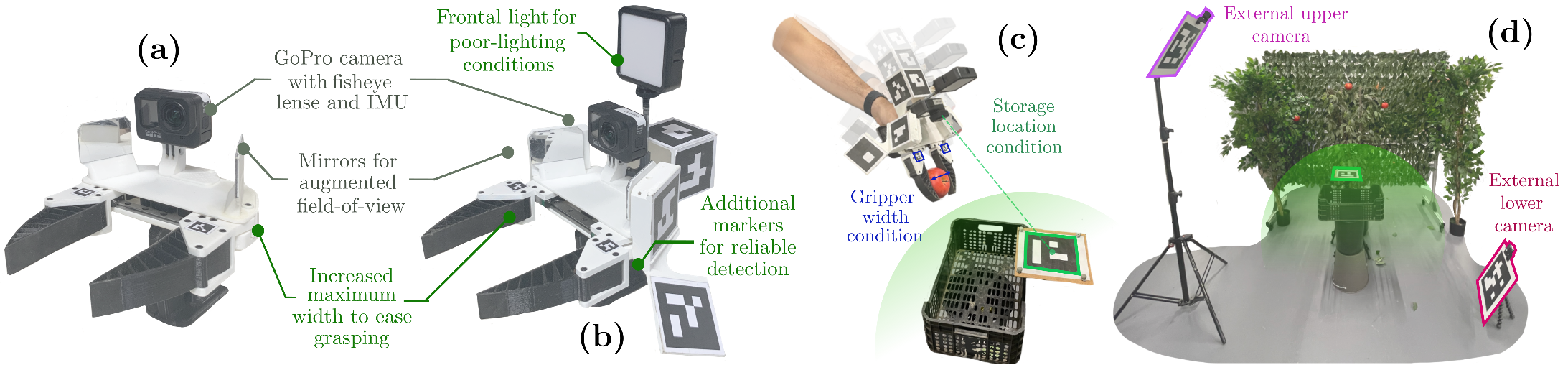}
    \hspace{-3mm}
    \caption{\centering Standard (a) and proposed modified version for agricultural settings (b) of the UMI hand-held grippers (performed modifications are highlighted in green), together with the proposed setups for sample extraction solution (c) and EKF-based generation of trajectories (d).}
    \label{fig:setup_gripper_comparison}
    \vspace{-5mm}
\end{figure}

Customary LfD techniques on agricultural tasks are challenging to be introduced as physically hand-guiding is required by expert users. Nevertheless, recent techniques are heading towards democratizing sample acquisition through the usage of intuitive hand-held grippers that mimic robot ones. Among the existing solutions in this direction, this work proposes to make use of the Universal Manipulator Interface (UMI) framework~\cite{chi_universal_2024} to retrieve human demonstrations of the complex manipulation skills required in fruit harvesting, and transferring them to robotic manipulators. The main advantages of the UMI framework are (i) a low-cost 3D printed hand-held gripper embedding mirrors that, together with the usage of a fisheye camera lense, enlarge the field-of-view of the scene (Figure~\ref{fig:setup_gripper_comparison}a), and (ii) a policy interface based on diffusion policies for robust skill learning from a small number of samples. Nevertheless, regarding the demo acquisition phase, the proposed pipeline has two main drawbacks for its application on agricultural settings:
\begin{enumerate}
    \item Human demonstrations are extracted individually by recording one video per each task sample, as in considered scenarios only a single task is meant to be performed, e.g. placing a cup on top of a plate. In fruit harvesting, the same task is repeated multiple times on the same scene, such that no restart is required, and individual sample extraction would increase idles times and lead to user-induced errors.
    \item To generate the associated trajectories for each task sample, UMI relies on the visual-inertial Simultaneous Localization and Mapping (SLAM) technique provided by ORBSLAM3~\cite{campos_orb-slam3_2021}. Thus, there is an initial mapping phase using a video that needs to contain enough scene information to extract the associated map, such that afterwards all the task samples can then be localized on it. As stated by UMI authors, mapping phase highly constraints the generation of samples, and ORBSLAM3 is not reliable enough for scenarios with texture-deficient environments, such as outdoors ones that might be found in agricultural settings. 
\end{enumerate}
This paper presents solutions to tackle both:
\begin{enumerate}
    \item An algorithm to extract all task samples from a single video containing all of them, using task events detected through marker detection and grasping monitoring, as described in Section~\ref{sec:demo_extract}.
    \item An Extended Kalman Filter (EKF)~\cite{julier_ekf_2004} implementation to generate the trajectories of the extracted task samples, using the on-board camera inertial measurements and the localization of additional markers added to the original hand-held gripper from an external multi-camera system, as shown in Fig.~\ref{fig:setup_gripper_comparison}b, whose description and required setup are detailed in Section~\ref{sec:localization}.
\end{enumerate}
Additional modifications have been performed on the hand-held gripper to better suit the agricultural setting, also highlighted in Fig.~\ref{fig:setup_gripper_comparison}b. To assess the proposed approach, results are provided on an apple harvesting task in Section~\ref{sec:results}.

\vspace{-2mm}
\section{Task Sample Extraction}~\label{sec:demo_extract}
\vspace{-5mm}

Many fruits can be picked up within the same scene, following all the operations the same steps: identification (from a fixed position), approaching, grasping. ``detaching'' and placing in a storage location, i.e.~a box or a conveyor belt. Thus, in a continuous demonstration with multiple picking operations, samples can be differentiated by identifying the last placing step. The approach presented in this paper uses two task events features to make this identification: gripper width and proximity to placing location. 

UMI hand-held gripper feature ArUco~\cite{GARRIDOJURADO2016481} markers on each finger, such that gripper width can be used as an action for the diffusion policy learning process. This pipeline has been reused to determine when the gripper grasps and releases the fruit and hence when the picking up task has finished. A condition using this metric can be defined from the maximum gripper width, i.e.~open position, and the maximum allowed fruit size, which can be provided by a human expert. 

Gripper-width condition suffices to identify the end of each picking task, but considering it alone sets the beginning of the next task just after the previous release of the fruit. Using these samples for the posterior diffusion policy learning phase could lead the robot to seek for the placing phase as a staring point before performing next picking operation. To mitigate this effect, an additional condition is introduced, based on the detection of an ArUco marker placed in a fixed position close to the storage location. Hence, the condition can be defined from the distance to the marker, such that it allows to identify when the gripper ``leaves'' the storage location, according to a tuned threshold. The proposed solution merges both conditions such that, once the fruit is released, the second condition is monitored to set the end of the current task and the beginning of the next one. The setup required for this solution is shown in Fig.~\ref{fig:setup_gripper_comparison}c.

\vspace{-2mm}
\section{EKF-based generation of trajectories}~\label{sec:localization}
\vspace{-5mm}

As aforementioned, the proposed solution for trajectory generation adds to the default UMI setup two external cameras for the detection of the ArUco markers placed on the hand-held gripper. Hence, using the known pose of these markers within the gripper, an estimation of the hand-held gripper pose can be obtained per marker detection. The solution presented in this paper considers both the detection of the unused markers in a cube arrangement from the default version and new ones placed on additional lateral pieces, as shown in Fig.~\ref{fig:setup_gripper_comparison}b. These new markers enable the localization of the hand-held gripper from many different external points and also for troublesome movements such as the rotations around the forward direction axis, which are usually performed for the ``twist'' movements needed to detach the fruit from the branch. External camera location also serves this purpose: first one is placed to get a view from above, and second one on the opposite side to get a view from below, as shown in  Fig.~\ref{fig:setup_gripper_comparison}d. 

For the generation of task sample trajectories, a common reference frame for marker localization needs to be defined. Each camera is mounted at a fixed position above an AprilTag Marker~\cite{olson2011tags}, such that their mutual localization can be used to minimise their relative position and orientation error, ensuring that the hand-held gripper pose estimations are coherent between both cameras. To handle occlusions or fast-movements that might hinder marker detection, the solution presented in this paper proposes to combine the marker-based localizations of the hand-held gripper with the inertial measurements from the on-board camera, using an EKF implementation. Generally speaking, this solution aims at improving the reliability of the generation of task sample trajectories by substituting the on-board visual localization feature from ORBSLAM3 by a marker-based localization from an external multi-camera system, but still taking advantage from inertial measurements for robustness and accuracy. 


%
\vspace{-3mm}
\section{Results}~\label{sec:results}
\vspace{-7mm}


Experiments have been performed on the realistic apple orchard in a wall configuration depicted in Fig.~\ref{fig:setup_gripper_comparison}d. The external cameras 
together with the additional 3D printed parts and the on-board frontal and external lights, adds 600~\euro \hspace{1mm} to the UMI setup, maintaining the budget under 900~\euro. In terms of software, intrinsic camera and IMU parameters are calibrated using the Open IMU-Camera Calibrator library~\footnote{Urban S., OpenICC - \url{github.com/urbste/OpenImuCameraCalibrator}}, the minimisation of external camera pose error is performed using Nelder-Mead method, and used EKF implementation is provided by a customary robot localization library~\footnote{C.R. Analytics, Robot Loc. for ROS - \url{github.com/cra-ros-pkg/robot_localization}}. It should be noted that default ORBSLAM3 configuration from UMI pipeline fails to reliable generate task trajectories and the number of visual features had to be increased from 2500 to 3000.




\begin{figure}[!t]
    \centering
    \includegraphics[width=0.97\textwidth]{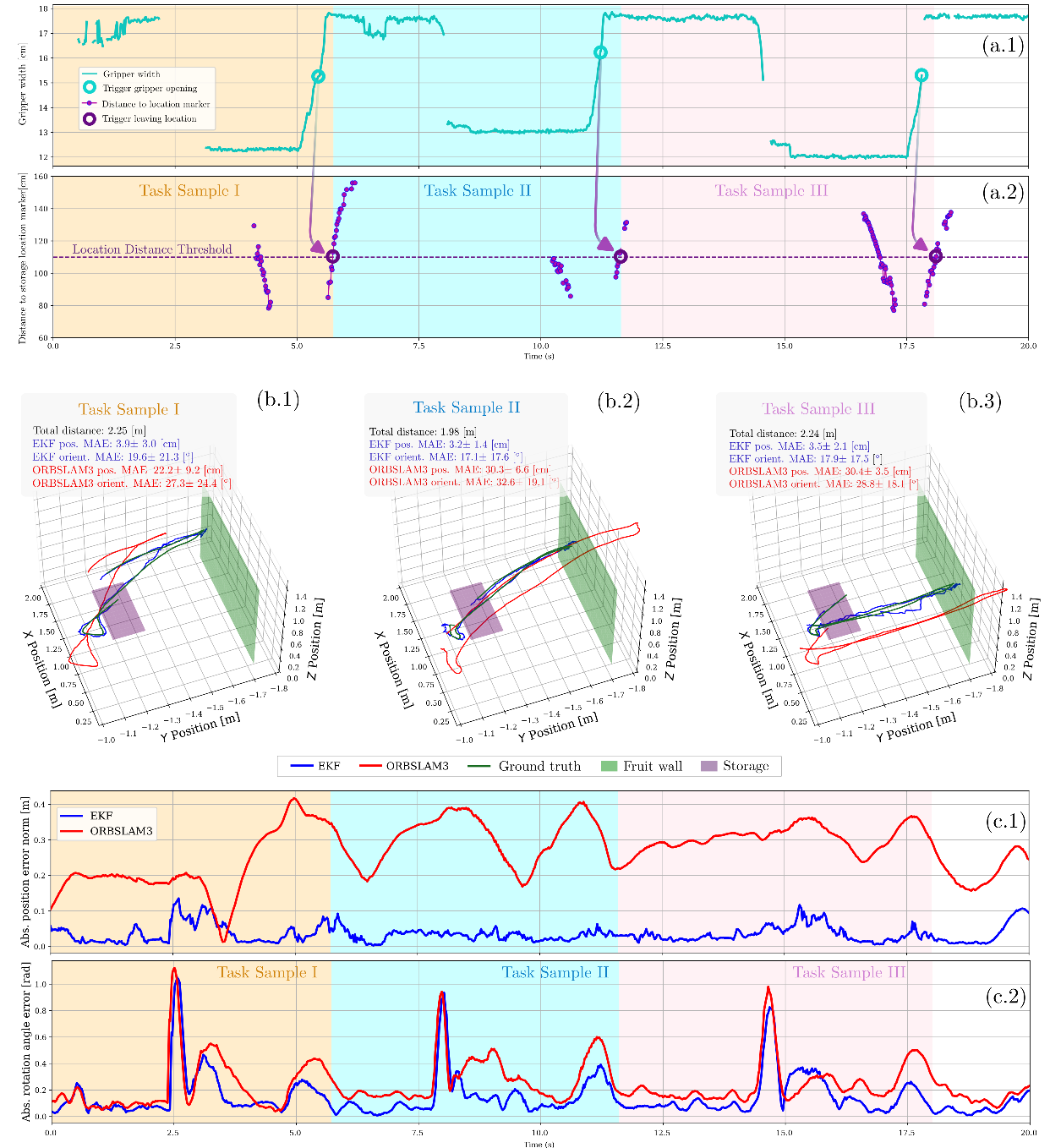}
    \vspace{-5mm}
    \caption{\centering Results on task extraction according to task-event criteria (a) and the generated trajectories together with position and orientation errors (b) and their evolution (c) for both ORBSLAM3 and EKF implementation w.r.t. ground truth.}
    \label{fig:results}
    \vspace{-6mm}
\end{figure}

Results have been obtained for a complete demonstration containing the harvesting of three apples within the scene, including approaching, grasping,``de\-tach\-ing'' movement and placing on a storage location. Figs.~\ref{fig:results}a shows the monitoring signals for gripper width and distance to storage location marker, together with the thresholds defined for the setup configuration that set the transition between tasks. As it can be seen, after opening trigger, monitoring location distance correctly sets the end of each task. The trajectories generated for each tasks are shown in Figs.~\ref{fig:results}b, using both ORBSLAM3 and proposed EKF implementation, together with the ground truth obtained from an Optitrack motion capture system~\footnote{Natural Point, Optitrack: Motion Capture System - \shortlink{optitrack.com/}}. Using this data, position and orientation errors of each method have been obtained and included for each task, and their time-evolution has been graphically depicted in Figs.~\ref{fig:results}c. This outcome evidences the improvement in terms of accuracy achieved by the proposed EKF implementation, which achieves lowest errors for all three tasks. ORBSLAM3 does deliver trajectories that set greater covered distances between consecutive points that visually render "oversized" trajectories w.r.t. ground truth ones. This could be due to the performance degradation from the visual-based localization component, that might be amplified from the usage of a non-linear lens type. On the other hand, orientation errors of ORBSLAM3 remain closer to EKF ones, shown both ``peaks'' in their temporal evolution (Figure~\ref{fig:results}c) when performing the ``detaching'' movement, which requires a fast rotation of the hand-held gripper. A video for the whole demonstration and the performance of task extraction and ORBSLAM3 and EKF trajectory generation can be found in the following link:~\url{https://youtu.be/BaHs9nDfU18?si=5nMPt-nHDN1iZZwt}






\vspace{-3mm}
\section{Conclusions and future work}~\label{sec:conclusions}
\vspace{-6mm}

This work focuses on the acquisition of task samples on agricultural settings using the UMI hand-held gripper, by (i) introducing a task sample extraction algorithm from a continuous demonstration and (ii) substituting the trajectory generation pipeline based on ORBSLAM3 by an EKF implementation that combines on-board inertial measurements and marker-based localization from an external multicamera system. Experimental results are obtained for a continuous harvesting task on a realistic fruit wall setting, showing that tasks are successfully extracted and proposed EKF implementation outperforms ORBSLAM3 in position and orientation accuracy. In future works, the implications of this approach on diffusion policy learning should be addressed, together with adapted configurations of EKF implementations, e.g. for high-speed movements.

%

%
\vspace{-3mm}
\section*{Acknowledgements}
\vspace{-2mm}

This work was financially supported by the Catalan Government through ACCIO-Eurecat (Project Replicate) funding grant, and AgRibot project through European Union’s Horizon 2024 research and innovation programme under GA No. 101183158. Views and opinions expressed are however those of the author(s) only and do not necessarily reflect those of the European Union or European Innovation Council and SMEs Executive Agency (EISMEA). Neither the European Union nor the granting authority can be held responsible for them.

\vspace{-4mm}
%
%
\bibliographystyle{styles/bibtex/splncs03_unsrt.bst}
\bibliography{library.bib}

\end{document}